\documentclass{Interspeech}

\interspeechcameraready

\title{Developing a High-performance Framework for Speech Emotion Recognition in Naturalistic Conditions Challenge for Emotional Attribute Prediction}

\author[affiliation={}, equalcontribution]{Thanathai}{Lertpetchpun}
\author[affiliation={}, equalcontribution]{Tiantian}{Feng}
\author[affiliation={}]{Dani}{Byrd}
\author[affiliation={}]{Shrikanth}{Narayanan}

\affiliation[nocounter]{University of Southern California}{Los Angeles}{USA}
\email{lertpetc@usc.edu, tiantiaf@usc.edu}

\usepackage{comment}
\usepackage{multirow}
\usepackage{xcolor}
\usepackage{booktabs}
\usepackage{hyperref}
\begin{document}

\maketitle

\begin{abstract}
Speech emotion recognition (SER) in naturalistic conditions presents a significant challenge for the speech processing community. Challenges include disagreement in labeling among annotators and imbalanced data distributions. This paper presents a reproducible framework that achieves superior (top 1) performance in the Emotion Recognition in Naturalistic Conditions Challenge (IS25-SER Challenge) - Task 2, evaluated on the MSP-Podcast dataset. Our system is designed to tackle the aforementioned challenges through multimodal learning, multi-task learning, and imbalanced data handling. Specifically, our best system is trained by adding text embeddings, predicting gender, and including ``Other'' (O) and ``No Agreement'' (X) samples in the training set. Our system's results secured both first and second places in the IS25-SER Challenge, and the top performance was achieved by a simple two-system ensemble.
\end{abstract}

\section{Introduction}
Recent advances in foundation models have opened new frontiers in speech processing, leading to more accurate and robust solutions in automatic speech recognition \cite{shi23g_interspeech, shi24g_interspeech}, speaker recognition \cite{peng24_interspeech}, and spoken language understanding \cite{futami24_interspeech}. Despite the success of these foundation models, accurately predicting emotions expressed in speech in natural interaction contexts still encounters several challenges, including perception (annotator) subjectivity  \cite{chien2024balancing,booth2024people}, imbalanced distribution in emotion labels, and mismatch in speaker and environment settings \cite{LeeSpeechEmo-ProcIEEE2023}. The Speech Emotion Recognition in Naturalistic Conditions Challenge (IS25-SER Challenge) offers a platform to investigate some of these limitations within a research community ``challenge" setting.

In this paper, we focus on Task 2 IS25-SER Challenge~\footnote{\href{https://lab-msp.com/MSP-Podcast_Competition/IS2025/}{https://lab-msp.com/MSP-Podcast\_Competition/IS2025/}}~\cite{Naini_2025}. Task 2 involves predicting the level of three-dimensional motion attributes: arousal, valence, and dominance \cite{Grimm2007Primitives-basedevaluationandestimations}. Arousal reflects the intensity of the emotion, while valence indicates its polarity. Dominance represents the level of control a person feels in an emotional state. These emotion attributes have proved challenging to approach computationally; this was reflected in the previous Odyssey-Speech Emotion Challenge \cite{goncalves2024odyssey}, in which none of the participating teams scored higher than 0.5327 in concordance correlation coefficient (CCC) \cite{atmaja2021evaluation}. Consequently, in this study, we follow that baseline to extract speech embeddings, namely a pre-trained speech model (WavLM Large \cite{chen2022wavlm}) with a series of linear classification layers.

Aside from the architecture choices, we aim for a simple and reproducible system design. Although ensembling a set of trained models is a common practice that yields competitive performance in several domains~\cite{tran24_asvspoof, tomilov21_asvspoof, thienpondt21_interspeech}, here we demonstrate that a combination of just two systems can achieve promising performance—top-tier in the challenge, outscoring the baselines by a considerable margin. Specifically, our best system reaches an average CCC of $0.6076$, $4.8\%$ higher than the baseline. While this improvement may seem small overall, it is 35.43\% higher than the first runner-up. While our current system yields state-of-the-art performance, our exploration during the challenge reveals many design considerations that lead to compromises in performance. Therefore, we share both positive findings, and the limitations that remain, as insights for future system design.

While previous work by \cite{lertpetchpun2023instance} suggests that speaker information is beneficial in predicting emotions, most systems in the recent Odyssey-SER challenge do not integrate speaker-related information, such as gender or speaker identity. This motivates us to explore joint modeling of the emotional attributes and speaker information. Specifically, we experiment with speaker representation learning and show that not all speaker-related information is helpful for emotion attribute prediction. Our results suggest that learning gender information improves the model on the SER task, whereas learning speaker representations leads to decreased performance. Hence, we perform a detailed analysis to explain why speaker representation learning fails to improve SER performance in this challenge.

Furthermore, we address the challenge of imbalanced emotion classes in the training dataset by undersampling, so that the model identifies emotion attributes more uniformly across each primary emotion. Our results show that this simple method is effective in improving SER performance. Building upon this, we further investigate the impact of including or excluding speech samples with mixed annotations, labeled as ``Other" (O) and ``No Agreement" (X). Most studies choose to discard these labels without an explicit rationale. Instead, our findings indicate that these labels contain valuable information that helps the model better learn emotion attributes.

Our model is at \href{https://github.com/tiantiaf0627/vox-profile-release}{https://github.com/tiantiaf0627/vox-profile-release}. We believe that our system can serve as a new baseline candidate for future SER challenges. The key contributions and findings are summarized below:

\begin{itemize}[leftmargin=*]
    \item Incorporating text representations alongside speech embeddings significantly improves model performance.
        
    \item Leveraging auxiliary labels such as gender information during training enhances the model’s ability to generalize.
        
    \item Addressing data imbalance through undersampling improves model performance, and accelerates training speed.
    
    \item Including ``Other" and ``No Agreement" labels in training, unlike most previous work, improves SER performance.
\end{itemize}

\section{Method}
\begin{figure}
    \centering
    \includegraphics[width=1\linewidth]{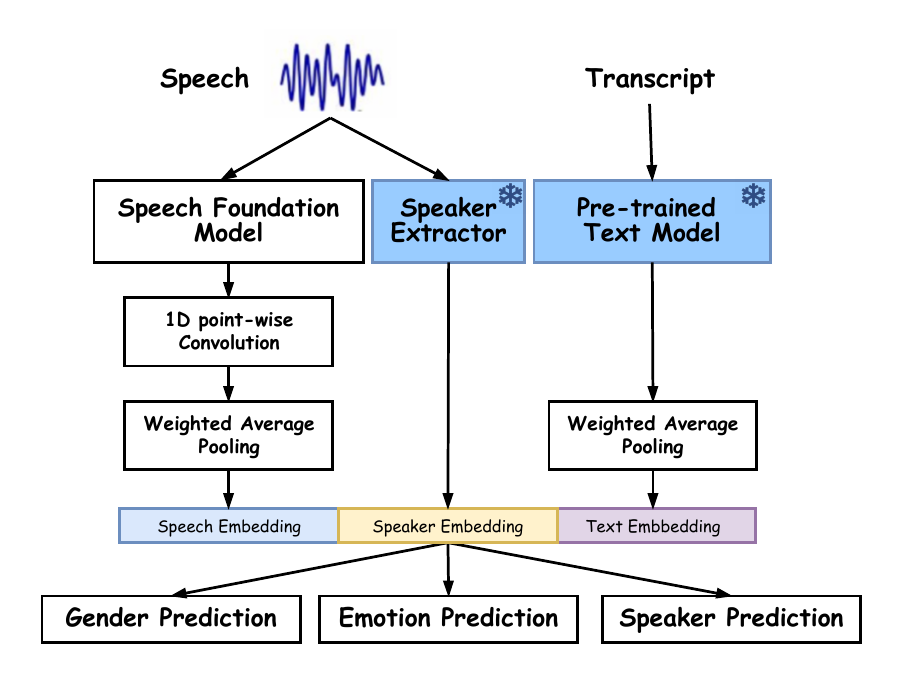}
    \caption{Our proposed multimodal SER framework utilizes a speech foundation model, speaker extractor, and text model. The pre-trained speaker extractor and pre-trained text model are frozen in every setting of our framework.}
    \label{fig:framework}
    \vspace{-3mm}
\end{figure}

\subsection{Foundation Models}
Foundation models such as WavLM \cite{chen2022wavlm} and Whisper \cite{radford2023robust} have been shown to improve the performance in SER  \cite{pepino21_interspeech, feng2024foundation}. In the previous Odyssey-SER Challenge, the baseline system \cite{goncalves2024odyssey}, which used the foundation model following downstream MLPs, outperformed every submission. Thus, following the baseline system, our system is built upon speech foundation models, including WavLM and Whisper. For WavLM, we computed the encoder output using a weighted average pooling from all hidden outputs from encoder layers. On the other hand, for Whisper. we used the hidden output from the last layer as the encoder output Specifically, we process the encoder outputs using 1D point-wise convolutions.

Furthermore, previous studies find that text embeddings from a pre-trained text model \cite{feng2024foundation, li2022fusing} and speaker embeddings from a speaker feature extractor \cite{lertpetchpun2023instance} are beneficial to improve SER. We concatenated the speech embedding, text embedding, and speaker embedding to obtain the final representation. Our proposed framework is illustrated in Figure~\ref{fig:framework}.

\subsection{Attribute Representation Learning}
The final representations are fed into three distinct 2-layer MLPs to predict arousal, valence, and dominance in a regression task. Since the challenge uses the concordance correlation coefficient (CCC) as the evaluation metric \cite{atmaja2021evaluation}, we apply CCC loss as our learning objective. Moreover, we perform multi-task learning to simultaneously learn emotion attributes alongside speaker-specific information such as gender and speaker identity. To achieve this, we pass the final representation through a separate 2-layer downstream model to predict the gender information, optimizing with cross-entropy loss. Subsequently, we feed the same representation into another downstream model to learn the speaker identity representation. When training the model for speaker prediction, we use additive angular margin loss \cite{deng2019arcface}, a widely used loss for speaker verification tasks, and exclude the speaker embedding from the final representation. Our final loss function is as Eq.~\ref{eq:finalloss}.

\vspace{-2.5ex}
\begin{equation}
    L = L_{emotion} + \alpha L_{gender} + \beta L_{speaker}
    \label{eq:finalloss}
\end{equation}
\vspace*{-2.5ex}

where $L_{emotion}$ represents summation of the CCC loss for each attribute, $L_{gender}$ indicates the cross-entropy loss between for gender prediction, $L_{speaker}$ represents the AAM-Softmax loss for speaker prediction, and $\alpha$ and $\beta$ denote weights of each loss. 


\subsection{`Other' and `No Agreement' Label Inclusion}
One debated topic in SER is how to handle No Agreement (X) samples and Other (O) samples (we will refer to these two labels as ``O/X''). No Agreement (X) samples represent speech samples that lack consensus in emotion annotation, while the Other (O) category indicates that the emotion category falls outside the eight primary emotions described in \cite{goncalves2024odyssey}. Many previous studies simply excluded these two labels from training, development, and test \cite{harm24_odyssey, chen24_odyssey}. However, the speech samples of these two labels account for 22.46\% of the training set. Instead of discarding these samples, we include them in the training set and demonstrate their effectiveness in improving SER modeling.

\subsection{Engineering Design Choices to Improve Performance }

Despite our efforts to tackle the challenges described in the previous section, many hurdles still need to be addressed. As announced by the organizer, the test set is balanced in primary emotion labels. Therefore, the proportion of each primary emotion in the test set differs substantially from that in the training set. Training the model on the training set can lead the model to overfit to the majority class, such as Neutral. Moreover, validating our model to choose the best model based on the development may not reflect the actual score on the test set due to the distribution mismatch between the development and test set. 

\vspace{1.5mm}
\noindent \textbf{Validation Selection} 
We develop another auxiliary metric to choose our best model. We aim to establish another CCC to reflect on the balanced dataset using existing imbalanced datasets. Specifically, we perform $n$ trials to compute the average CCC in this manner. In each trial, we randomly select $k$ samples from each primary emotion and compute CCC using those samples. Although the metrics may exhibit noise due to randomness in the sample selection, this approach ensures that the CCC is evaluated on the balanced set following the same primary emotion distribution as the test set.

\vspace{1.5mm}
\noindent \textbf{Undersampling}
We apply the undersampling to ensure that each emotion sample is fed into the model with equal probability. We find $m = min_i(|emotion_i|)$ where $|\cdot|$ denotes the number of samples in that emotion. Then, we randomly select $m$ samples from each class into the training set.

\vspace{1.5mm}
\noindent \textbf{Target Normalization}
We apply the sigmoid function to the model output to map the emotion attribute prediction to the range $[0, 1]$. These values are subsequently mapped back to the range $[1, 7]$ using the min-max normalization. Specifically, the output can be computed as: 
\vspace{-2mm}
\begin{align}
    \hat x_i = x_i \times [max_j(y_{ij}) - min_j(y_{ij})] + min_j(y_{ij})
\end{align}

\vspace{-2mm}
where $x_i$ represents the prediction of each attribute and $y_{ij}$ denotes the value of each sample ($j$) from each attribute ($i$).

\section{Experimental Setup}
\subsection{MSP-Podcast Dataset}
The MSP-Poscast dataset v1.12 \cite{lotfian2017building} was used as the dataset for the IS25-SER Challenge. The dataset comprises podcast data from the Internet which are spontaneous speech and contain human emotions expressed in a naturalistic context. The dataset was annotated with different emotion categories and attributes through crowd-sourcing. Moreover, there are at least five annotators per sample to ensure the reliable labeling. However, as annotators sometimes cannot converge on a single emotion, such samples will be labeled as ``No Agreement'' (X) and are excluded from the test set. The dataset contains five subsets: the training set, the development set, and the three unique test sets. In the challenge, we were given the training and development set, while, the test-3 set, for which annotation labels were not made publicly available, was used for test evaluation. The annotators also labeled each sample with regard to three different attributes: arousal (negative to positive), valence (calm to active), and dominance (weak to strong), with ratings on a scale from 1 to 7. Our experiment used the entire training set, consisting of ten emotion classes: Neutral, Happy, Sad, Disgust, Angry, Contempt, Fear, Surprise, Other, and No Agreement. Detailed information about the training and development sets are reported in Table~\ref{tab:dataset}

Since the challenge rules allowed the training and validation set to be used arbitrarily, we chose to include some parts of the validation set in the training set to improve the system. Specifically, the samples with label ``Other" (O) and ``No Agreement" (X) were included in the training set and excluded from the development set. 

\begin{table}[h]
    \centering
    \vspace{-2mm}
    \caption{Dataset statistics of IS25-SER challenge.}
    \vspace{-2mm}
    \begin{tabular}{lccc}
        \toprule
         & \textbf{Training} & \textbf{Development} & \textbf{Test} \\
        \midrule
        \textbf{Neutral} & 29,243 & 7,423 & 400 \\
        \textbf{Happy} & 16,717 & 6,344 & 400 \\
        \textbf{Sad} & 6,306 & 2,341 & 400 \\
        \textbf{Disgust} & 1,432 & 542 & 400 \\
        \textbf{Angry} & 6,731 & 5,836 & 400 \\
        \textbf{Contempt} & 2,495 & 1,459 & 400 \\
        \textbf{Fear} & 1,120 & 326 & 400 \\
        \textbf{Surprise} & 1,120 & 987 & 400 \\
        \textbf{Other} & 2,948 & 642 & 0 \\
        \textbf{No Agreement} & 15,932 & 6,061 & 0 \\
        \bottomrule
    \end{tabular}
    \vspace{2mm}
    \label{tab:dataset}
    \vspace{-5mm}
\end{table}
\subsection{Experimental Details}
Although we acknowledge that hyperparameter tuning for each different system can help improve the aggregate models significantly, we did not focus on this aspect for this challenge: —our goal in this work was to design a reproducible and simple state-of-the-art pipeline. Consequently, all the systems were trained with the same hyperparameters. We used a learning rate of 5e-5 and trained the model for a total of 50 epochs. The filter size in the downstream model is 256 across all experiments. The maximum input speech duration is 15 seconds. The best model was selected based on the performance on the validation set. For the multitask learning, we used $\alpha=1$ and $\beta=0.1$. 

The systems are trained using the undersampling method as the default method in most experiments. We use WavLM Large and Whisper Large-V3 for the speech foundation model, and RoBERTa-Large as the pre-trained text model. Following the challenge baseline, we fine-tune the pre-trained WavLM Large along with the downstream models. However, we froze the model for the Whisper Large-V3 and RoBERTa-Large. We will add the prefix ``MM (multimodal)'' to the model to indicate that the model uses text embedding as an auxiliary input. As we did not have access to the test set and we can only submit the score to the challenge once a week, it was challenging for us to do a systematic evaluation on the test set. Therefore, the performances except for Table~\ref{table:main_results} are reported on the validation set.

\section{Results and Discussion}

\subsection{Pre-trained Text Model}
First, we investigated whether textual information impacts the SER performance by adding text embedding to the final representation. We designed an experiment to compare WavLM Large and WavLM Large + RoBERTa-Large (MMWavLM). Table~\ref{tab:text} shows that the MMWavLM performs better in every attribute prediction. 

\begin{table}[t]
    \centering
    \caption{Comparison between models trained without text embeddings and with text embeddings.}    
    \begin{tabular}{lccccc}
        \toprule
         & A & V & D & Average\\
        \cmidrule(lr){1-1} \cmidrule(lr){2-5}
        \textbf{WavLM} & 0.6530 & 0.7000 & 0.5651 & 0.6394 \\
        \textbf{MMWavLM} & \textbf{0.6694} & \textbf{0.7334} & \textbf{0.5735} & \textbf{0.6588} \\
        \bottomrule
    \end{tabular}
    \label{tab:text}
    \vspace{-6mm}
\end{table}

\subsection{Attributes Learning}
\begin{figure}[h]
    \centering
    \includegraphics[width=\linewidth]{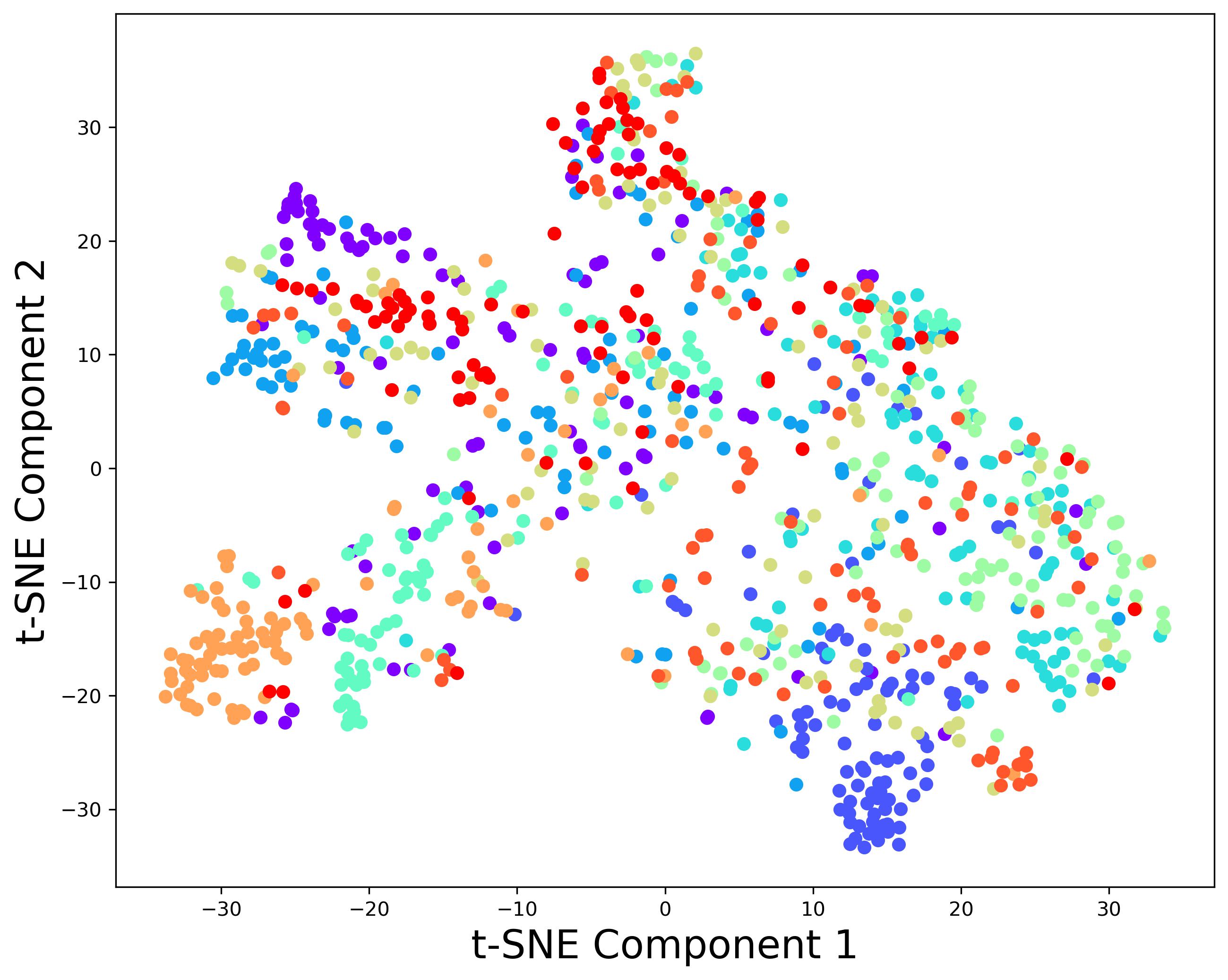}
    \caption{The t-sne plot of speaker embeddings from the pre-trained speaker extractor. Each color represents different speakers.}
    \label{fig:tse}
    \vspace{-3mm}
\end{figure}

As gender prediction has proven effective in classifying emotions \cite{gao23d_interspeech}, we investigate whether it can also improve the regression task. Table~\ref{tab:attribute} shows that multi-task learning with gender information increases the CCC scores in arousal, valence, and dominance predictions. However, adding the speaker embedding or predicting the speaker label worsens the model performance. We analyzed the speaker label to understand the reasons behind this. First, approximately 80\% of the shared dataset were represented by just 26\% of the total speakers, causing the model to be overfit to a limited subset of seen speakers. Second, the pre-trained speaker extractor does not generalize well on the dataset. We randomly selected 100 utterances from 10 speakers and extracted speaker embedding from them. As shown in Fig~\ref{fig:tse}, speaker embeddings failed to cluster together, making it difficult for the model to learn.

\begin{table}[h]
    \centering
    \caption{Comparison of models trained with speaker embeddings, speaker prediction, and gender prediction.}
    \vspace{-2mm}
    \begin{tabular}{lccccc}
        \toprule
         & A & V & D & Average\\
        \cmidrule(lr){1-1} \cmidrule(lr){2-5}
        \textbf{MMWavLM} & \textbf{0.6694} & 0.7334 & 0.5735 & 0.6588 \\
        \cmidrule(lr){1-1} \cmidrule(lr){2-5}
        \textbf{+ S Emb} & 0.6483 & 0.6962 & 0.5643 & 0.6363 \\
        \textbf{+ S Pred} & 0.6428 & 0.7141 & 0.5684 & 0.6418 \\
        \cmidrule(lr){1-1} \cmidrule(lr){2-5}
        \textbf{+ G Pred} & 0.6662 & \textbf{0.7367} & \textbf{0.5787} & \textbf{0.6605} \\
        \bottomrule
    \end{tabular}
    \label{tab:attribute}
    \vspace{-7mm}
\end{table}

\subsection{Undersampling}
As shown in Table~\ref{tab:subsamplinng}, undersampling significantly helps the model learn the true distribution of each emotion and prevents the model from overfitting only on certain emotions. Specifically, the CCC on valence increases significantly up to 31.57\%, leading the average CCC increases up to almost 10\%. Furthermore, our experiments show that training with undersampling helps the model converge faster, consequently reducing the training time.

\begin{table}[h]
    \centering
    \caption{Comparison between models trained with balanced and imbalanced data samples}
    \vspace{-2mm}
    \begin{tabular}{lccccc}
        \toprule
         & A & V & D & Average\\
        \cmidrule(lr){1-1} \cmidrule(lr){2-5}
        \textbf{Whole dataset} & 0.6620 & 0.5574 & \textbf{0.5795} & 0.5996 \\
        \textbf{Undersampling} & \textbf{0.6694} & \textbf{0.7334} & 0.5735 & \textbf{0.6588} \\
        \bottomrule
    \end{tabular}
    \label{tab:subsamplinng}
    \vspace{-5mm}
\end{table}

\subsection{Other and No Agreement labels}
Although O/X samples are not in test sets, adding or removing them substantially impacts performance. Table~\ref{tab:o/x} shows that removing data from the training set, whether it belongs to O or X, degrades performance. The results also show that adding O/X (V) samples to the training set improves the performance. Moreover, adding O/X (V) samples and removing non-O/X (T) samples give the best performance for the SER system. This indicates that, despite the annotators' lack of agreement in classifying emotions for these samples, they still provide valuable information to the SER system. However, due to submission limitations, we did not submit the system trained with the whole training set + O/X (V) - Non-O/X (T) to the leaderboard, even though it achieved the best performance on the validation set. 

\vspace{-2mm}
\begin{table}[h]
    \centering
    \caption{Comparison between models trained with different amount of data. ``O/X'' denotes Other (O) and No Agreement (X), ``+'' denotes including the data, ``-'' denotes removing the data, and (T) and (V) indicate whether the samples are from the training or validation set}
    \begin{tabular}{lccccc}
        \toprule
         & A & V & D & Average\\
        \cmidrule(lr){1-1} \cmidrule(lr){2-5}
        Training Set & 0.6694 & 0.7334 & 0.5735 & 0.6588 \\
        \cmidrule(lr){1-1} \cmidrule(lr){2-5}
        \textbf{- O/X(T)} & 0.6622	& 0.7342 & 0.5774 & 0.6579 \\   
        \textbf{- Non-O/X (T)} & 0.6692 & 0.7332 & 0.5720 & 0.6581 \\ 
        \cmidrule(lr){1-1} \cmidrule(lr){2-5}
        \textbf{+ O/X(V)} & 0.6655 & 0.7125 & 0.6034 & 0.6605 \\    
        \hspace{3mm} \textbf{- Non-O/X (T)} & \textbf{0.6830} & \textbf{0.7582} & \textbf{0.6145} & \textbf{0.6852} \\   
        \bottomrule
    \end{tabular}
    \label{tab:o/x}
    \vspace{-5mm}
\end{table}

\vspace{-2mm}
\subsection{Results of the IS25-SER Challenge}
The results from the IS25 SER Challenge leaderboard show that our two systems achieved first and second place in the Challenge. ``+ S Emb'' denotes that the model uses speaker embedding from the pre-trained speaker extractor in the final representation while ``+ Undersampling'' denotes that the model is training with the undersampling method. 

As shown in Table~\ref{table:main_results}, MMWavLM performs significantly better than WavLM, indicating that textual embeddings also improve the performance in the test set. Moreover, adding speaker embedding causes the model to overfit on the speaker information, leading to degraded performance, as suggested by our experiments on the validation set. 

Our best system consists of only two models: WavLM with gender prediction, undersampling technique and O/X (V) inclusion, and MMWhisper with only undersampling. We ensembled the system by averaging the CCC prediction from each one. The results indicate that ensembling only a few systems can exhibit strong performance. Moreover, we test a 3-system ensemble: WavLM with gender prediction, undersampling technique and O/X (V) inclusion; WavLM with gender prediction and undersampling technique; and MMWhisper with gender prediction, undersampling technique and O/X (V) inclusion. However, this 3-system underperforms the 2-system on the test set. Lastly, it is worth noting that since the number of submissions is limited, it is challenging to demonstrate which aspects of our methods contribute the most to improving scores. 

\begin{table}[t]
    \centering
    \caption{The results from the IS25-SER Challenge on Task 2. ``A'', ``V'', and ``D'' denote arousal, valence, and dominance, respectively.}
    \begin{tabular}{lccccc}
        \hline
        \toprule
        & A & V & D & Average \\
        \cmidrule(lr){1-1} \cmidrule(lr){2-5} 
        Baseline & 0.6385 & 0.6232 & 0.4775 & 0.5797 \\
        \cmidrule(lr){1-1} \cmidrule(lr){2-5} 
        \textbf{WavLM} & 0.5339 & 0.5992 & 0.4132 & 0.5155 \\ 
        \cmidrule(lr){1-1} \cmidrule(lr){2-5} 
        \textbf{MMWavLM} &&&& \\ 
        \textbf{+ Undersampling} & 0.6099 & 0.6189 & 0.4736 & 0.5675 \\    
        \hspace{3mm} \textbf{+ S Emb} & 0.6079 & 0.5934	& 0.4479 & 0.5497 \\ 
        \cmidrule(lr){1-1} \cmidrule(lr){2-5} 
        \textbf{3-system (2nd)} & 0.6794 & 0.634 & 0.4878 & 0.6004 \\
        \textbf{2-system (1st)} & \textbf{0.6829} & \textbf{0.642} & \textbf{0.4980} & \textbf{0.6076} \\
        \bottomrule
    \end{tabular}
    \vspace{2mm}
    \vspace{-6mm}
    \label{table:main_results}
\end{table}

\vspace{-2mm}
\subsection{On Further Improvement}
While our current system achieves competitive results in the Challenge, we identify several promising directions for developing next-generation, state-of-the-art SER systems. As suggested by the previous baseline in the Odyssey-SER Challenge \cite{goncalves2024odyssey}, training on a separate system for each attribute demonstrated strong performance. Moreover, instead of freezing the pre-trained Whisper Large-V3 and RoBERTa-Large, we could apply LoRa as proposed by \cite{harm24_odyssey, feng2023peft} to further improve the performance. We believe that integrating these two methods would not only improve the performance of our system but also maintain its simplicity.

\vspace{-2mm}
\section{Conclusion}
We proposed a simple and reproducible state-of-the-art model for the IS25 SER Challenge. Our system consists of the ensemble between pre-trained MMWavLM Large fine-tuned with gender prediction, inclusion of O/X label from both the training and validation sets, and using undersampling, with MMWhisper Large-V3 fine-tuned via undersampling. The ensemble system can gain performance up to 0.6076 in average CCC and achieved 1st place in the Challenge leaderboard. Lastly, our analysis shows why using learning speaker representation does not improve the performance of the SER system in this dataset.

\section{Acknowledgment}
We gratefully acknowledge support from IARPA ARTS (award number 140D0424C0067, JHU subcontract) from the Office of the Director of National Intelligence. NSF (SCH 2204942).

\bibliographystyle{IEEEtran}
\bibliography{mybib}

\end{document}